\documentclass[letterpaper, 10 pt, conference]{ieeeconf}  

\IEEEoverridecommandlockouts                              

\usepackage{graphics} 
\usepackage{epsfig} 
\usepackage{mathptmx} 
\usepackage{times} 
\usepackage{amsmath} 
\usepackage{amssymb}  
\usepackage[table]{xcolor}
\usepackage{booktabs}
\usepackage{makecell}
\usepackage{multirow}
\usepackage[T1]{fontenc}

\usepackage{multirow}
\usepackage{float}

\title{\LARGE \bf
Quadrotor Autonomous Landing on Moving Platform}

\author{Pengyu Wang$^{1}$, Chaoqun Wang$^{2}$ \\ Jiankun Wang$^{1}$, \textit{Senior Member, IEEE}, and Max Q.-H. Meng$^{3}$, \textit{Fellow, IEEE}
\thanks{*This work is partially supported by Shenzhen Key Laboratory of Robotics Perception and Intelligence (ZDSYS20200810171800001), Southern University of Science and Technology, Shenzhen 518055, China, and National Natural Science Foundation of China grant \#62103181, \emph{(Corresponding authors: Jiankun Wang, Max Q.-H. Meng).}}
\thanks{$^{1}$Pengyu Wang and Jiankun Wang are with Shenzhen Key Laboratory of Robotics Perception and Intelligence, and the Department of Electronic and Electrical Engineering, Southern University of Science and Technology, Shenzhen, China.
        {\tt\small 12250023@mail.sustech.edu.cn, wangjk@sustech.edu.cn}}%
\thanks{$^{2}$Chaoqun Wang is with the School of Control Science and Engineering, Shandong University, Shandong, China.
        {\tt\small chaoqunwang@sdu.edu.cn}}%
\thanks{$^{3}$Max Q.-H. Meng is with Shenzhen Key Laboratory of Robotics Perception and
Intelligence, and the Department of Electronic and Electrical Engineering, Southern University of Science and Technology, Shenzhen, China, on leave from the Department of
Electronic Engineering, The Chinese University of Hong Kong, Hong Kong, and also with the Shenzhen Research Institute of the Chinese University of Hong Kong in Shenzhen, China.
        {\tt\small max.meng@ieee.org}}%
}

\begin{document}

\maketitle
\thispagestyle{empty}
\pagestyle{empty}

\begin{abstract}

This paper introduces a quadrotor's autonomous take-off and landing system on a moving platform. The designed system addresses three challenging problems: fast pose estimation, restricted external localization, and effective obstacle avoidance. Specifically, first, we design a landing recognition and positioning system based on the AruCo marker to help the quadrotor quickly calculate the relative pose; second, we leverage a gradient-based local motion planner to generate collision-free reference trajectories rapidly for the quadrotor; third, we build an autonomous state machine that enables the quadrotor to complete its take-off, tracking and landing tasks in full autonomy; finally, we conduct experiments in simulated, real-world indoor and outdoor environments to verify the system's effectiveness and demonstrate its potential.

\end{abstract}

\section{INTRODUCTION}

In recent years, quadrotor Unmanned Aerial Vehicles (UAVs) that can take off and land vertically (VTOL) have played a key role in power line inspection, express delivery, and farmland sowing due to their flexibility and ease of deployment~\cite{gupte2012survey}\cite{zhu2017hawkeye}\cite{li2019coverage}. Among them, autonomous take-off and landing technology without external positioning (e.g. GPS and 
motion capture) is an important part, which is essential for UAVs to perform more complex tasks and cooperate with ground mobile robots.

Sani \emph{et al.}~\cite{sani2017automatic} proposed a visual and inertial navigation method combined with Kalman filter and proportional-integral-derivative (PID) controller to achieve relative pose estimation and accurate UAV landing on ground QR code. To estimate the state quantities of dynamic targets and track them efficiently, Lee \emph{et al.}~\cite{lee2012autonomous} adopted an image-based visual servoing method in two-dimensional space to generate reference velocity commands, which were fed to an adaptive sliding mode controller while considering an adaptive rule for the ground effect in landing process. However, these methods could only achieve static and very low-speed autonomous landing and at the same time only
indoor experimental verification.

Accurate position estimation of UAV and landing platform is necessary for autonomous landing. Mellinger \emph{et al.}~\cite{mellinger2010control} designed robust quadrotor planning and control algorithms for UAV's perching and landing, but they needed a VICON\footnote{https://www.vicon.com/} motion capture system to track the quadrotor as well as the landing surfaces. Daly \emph{et al.}~\cite{daly2015coordinated} proposed a coordinated landing control scheme. In their method, they used a joint decentralized controller to reach the rendezvous point for landing and the controller performed stably in the presence of delay. However, their approach relied on expensive Real-Time Kinematic (RTK) GPS on both UAV and Unmanned Ground Vehicle (UGV) for sub-centimeter positioning. The reliance on external positioning systems was not practical in many scenarios.

To let the UAV land on a fast-moving target, Borowczyk \emph{et al.} used Kalman filtering to calculate the accurate pose of the UAV relative to the landing pad by fusing multiple sources of measurement information. They used the proportional (P) controller in the distance and the proportional-derivative (PD) controller in the near, and the mobile platform could reach a high speed at 30km/h \cite{borowczyk2017autonomous}. However, the UAV did not perform motion planning when tracking and landing, which means that when there are obstacles on the way, it may cause collisions because the UAV cannot avoid obstacles.

To address the challenges mentioned above, the main contributions of this paper are summarized as follows:

(1) We design a landing positioning system based on AruCo squared fiducial marker~\cite{romero2018speeded} to be placed on the moving platform. The UAV can detect and calculate the pose relative to the UGV quickly, accurately and at a low cost.

(2) We utilize a gradient-based local motion planning method~\cite{zhou2020ego} to generate a collision-free, smooth and feasible reference trajectory for UAV in tracking and landing.

(3) We design an automatic state machine that enables UAV to achieve take-off, tracking and landing missions.

The rest of this paper is organized as follows. We first introduce the UAV-UGV system in Section II and then present the detection and positioning of the mobile landing platform in Section III. Furthermore, we demonstrate the local motion planning algorithm and landing state machine in Section IV. Finally, in Section V and Section VI, we conduct experiments, analyze the results and give future plans. 

\section{QUADROTOR SYSTEM DESIGN}

\begin{figure}[htb]
    \centering
    \includegraphics[width=0.8\columnwidth]{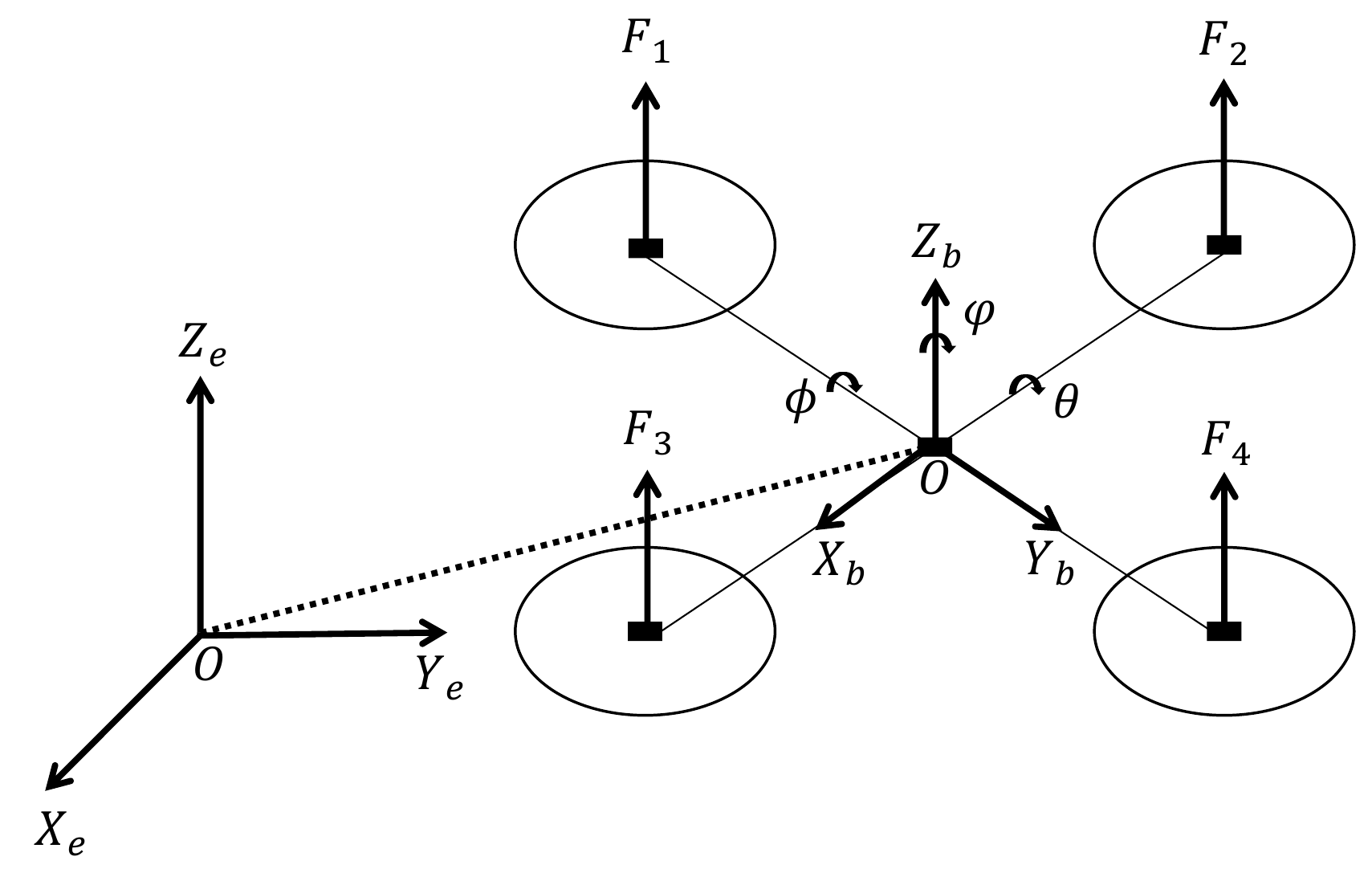}
    \caption{Coordinate systems of a quadrotor.}
    \label{fig:quadrotor_model}
\end{figure}
\vspace{-0.5cm} 

\subsection{Modeling}
In this paper, it is assumed that the quadrotor is a rigid body with uniform mass distribution and axisymmetric, the mass and rotational inertia are constant, and the center of gravity coincides with the geometric center, as shown in Fig.~\ref{fig:quadrotor_model}. $p=\left[ x\,\,y\,\,z \right]^T$ represents the position of quadrotor. $v=\left[ v_x\,\,v_y\,\,v_z \right] ^T$ and $\omega =\left[ \omega _x,\omega _y,\omega _z \right] ^T$ denote velocity and angular velocity, respectively. The roll/pitch/yaw angle of a quadrotor is represented by $\boldsymbol{\phi }=\left[ \varphi ,\,\,\theta ,\,\,\psi \right] ^T$. There are two coordinate systems: world inertial frame $E_q=\left\{ e_x,e_y,e_z \right\}$ and body-fixed frame $B_q=\left\{ b_x,b_y,b_z \right\} $. The transformation between two coordinate systems is represented by
\begin{equation}
\begin{aligned}
    & R_b = \\
    & \left[ \begin{matrix}
    	C\theta C\psi&		-C\phi S\psi +S\phi S\theta C\psi&		S\phi S\psi +C\phi S\theta C\psi\\
    	C\theta S\psi&		C\phi C\psi +S\phi S\theta S\psi&		-S\phi C\psi +C\phi S\theta S\psi\\
    	-S\theta&		S\phi C\theta&		C\phi C\theta\\
    \end{matrix} \right] 
\end{aligned}
\end{equation}

According to the Newton-Euler equation, the position and attitude dynamics equation of the quadrotor is expressed as follows \cite{pounds2006modelling}:
\begin{equation}
\begin{cases}
	\dot{v}_x=-\frac{T}{m}\left[ \left( sin\phi sin\psi +cos\phi sin\theta cos\psi \right) +F_{x}^{drag} \right]\\
	\dot{v}_y=-\frac{T}{m}\left[ \left( -sin\phi cos\psi +cos\phi sin\theta sin\psi \right) +F_{y}^{drag} \right]\\
	\dot{v}_z=-\frac{T}{m}\left[ cos\phi cos\theta +F_{z}^{drag}+mg \right]\\
\end{cases}
\end{equation}
\begin{equation}
\begin{cases}
	\dot{\omega}_x=\frac{1}{I_{xx}}\left[ \left( I_{yy}-I_{zz} \right) \omega _y\omega _z+\tau _{x}^{motor}+J_r\varOmega \omega _y \right]\\
	\dot{\omega}_y=\frac{1}{I_{yy}}\left[ \left( I_{zz}-I_{xx} \right) \omega _x\omega _z+\tau _{y}^{motor}-J_r\varOmega \omega _x \right]\\
	\dot{\omega}_z=\frac{1}{I_{zz}}\left[ \left( I_{xx}-I_{yy} \right) \omega _x\omega _y+\tau _{z}^{motor} \right]\\
\end{cases}
\end{equation}
where $T=c_T\varpi ^2$ is the thrust of the motor, $F^{drag}$ is air resistance, $I_i$ and $\tau _{i}^{motor}$ denote inertial and torque, respectively.

The position and attitude motion equation of the quadrotor can be written as
\begin{equation}
\dot{p}=v
\end{equation}
\begin{equation}
\left[ \begin{array}{c}
	\dot{\phi}\\
	\dot{\theta}\\
	\dot{\psi}\\
\end{array} \right] =\left[ \begin{matrix}
	1&		sin\phi tan\theta&		cos\phi tan\theta\\
	0&		cos\phi&		-sin\phi\\
	0&		sin\phi /cos\theta&		cos\phi /cos\theta\\
\end{matrix} \right] \left[ \begin{array}{c}
	\omega _x\\
	\omega _y\\
	\omega _z\\
\end{array} \right] 
\end{equation}

\subsection{Hardware Setup}
\begin{figure}[htb]
    \centering
    \includegraphics[width=0.4\columnwidth]{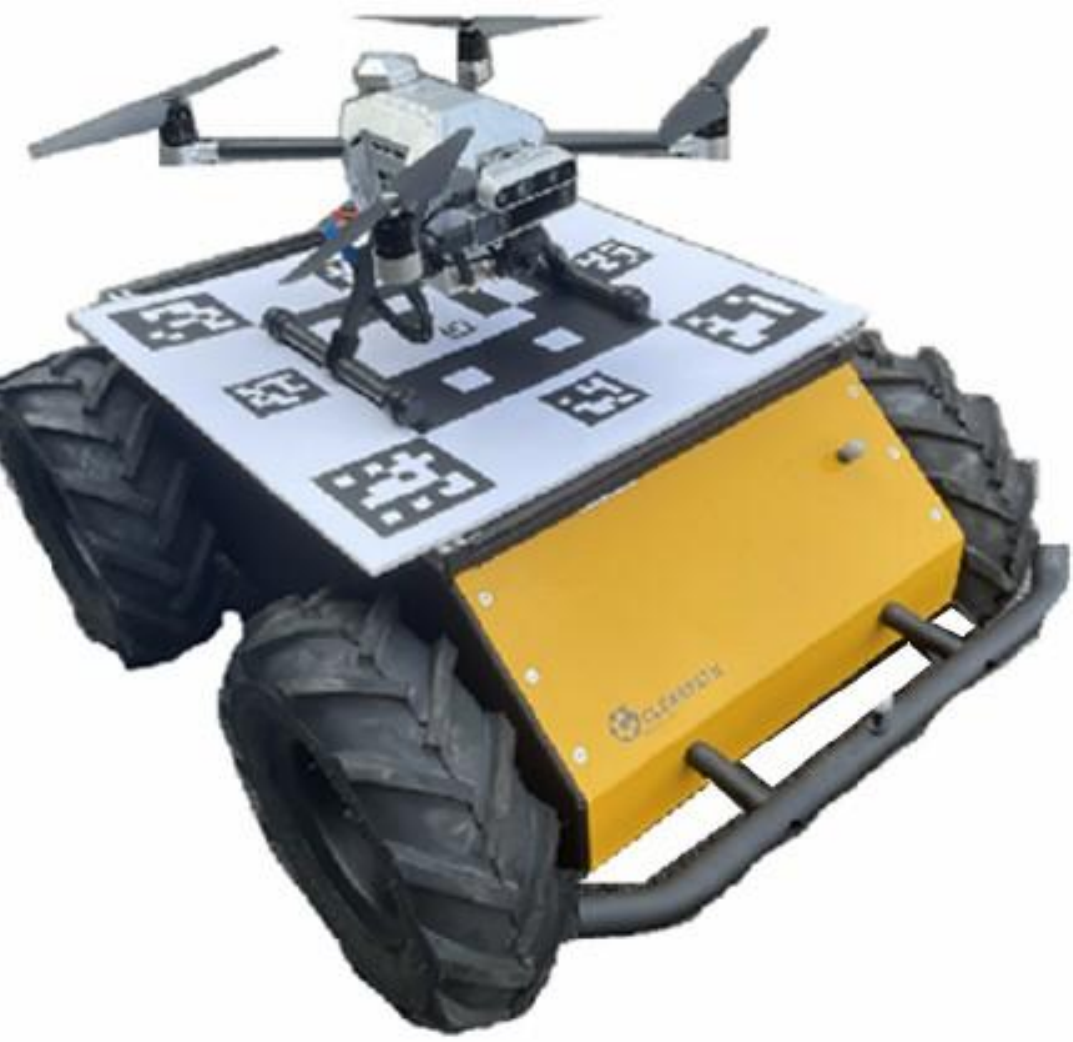}
    \caption{UAV-UGV system.}
    \label{fig:quadrotor}
\end{figure}

In this paper, we use a P450 quadrotor designed by Amovlab\footnote{https://wiki.amovlab.com/} and a Husky A200 UGV built by Clearpath\footnote{https://github.com/clearpathrobotics/clearpath\_husky}, as shown in Fig.~\ref{fig:quadrotor}. The core components of the quadrotor include open source autopilot hardware Pixhawk~\cite{meier2011pixhawk}, software PX4 flight control~\cite{meier2015px4}, onboard computer NVIDIA Jetson NX, monocular wide-angle camera, Intel Realsense T265 tracking camera and Intel Realsense D435i depth camera.  

\vspace{-0.5cm} 

\subsection{State Estimation}
\begin{figure}[htb]
    \centering
    \includegraphics[width=1.0\columnwidth]{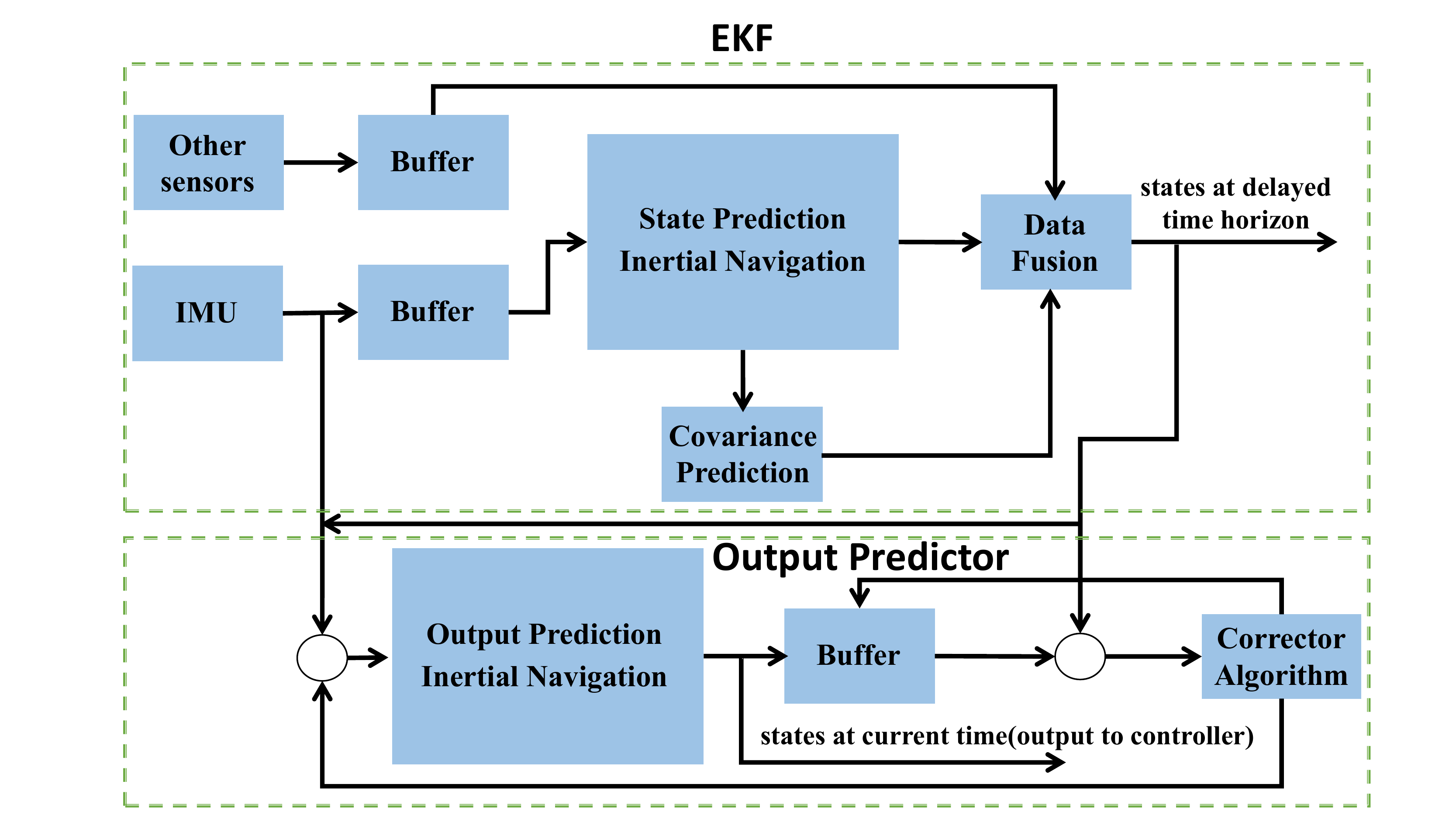}
    \caption{UAV state estimation.}
    \label{fig:state_estimation}
\end{figure}

PX4 provides an Extended Kalman Filter (EKF) \cite{ribeiro2004kalman} -based algorithm to process sensor measurements and calculate an estimate of flight states, as shown in Fig.~\ref{fig:state_estimation}. In order to get more accurate pose information and use it in a GPS-denied environment, we use the off-the-shelf Visual-Inertial Odometry (VIO)\footnote{https://github.com/IntelRealSense/realsense-ros} method in Intel Realsense T265 to obtain pose information of the UAV relative to the take-off point.

\vspace{-0.5cm} 

\subsection{Flight Controller}
\begin{figure}[htb]
    \centering
    \includegraphics[width=1.0\columnwidth]{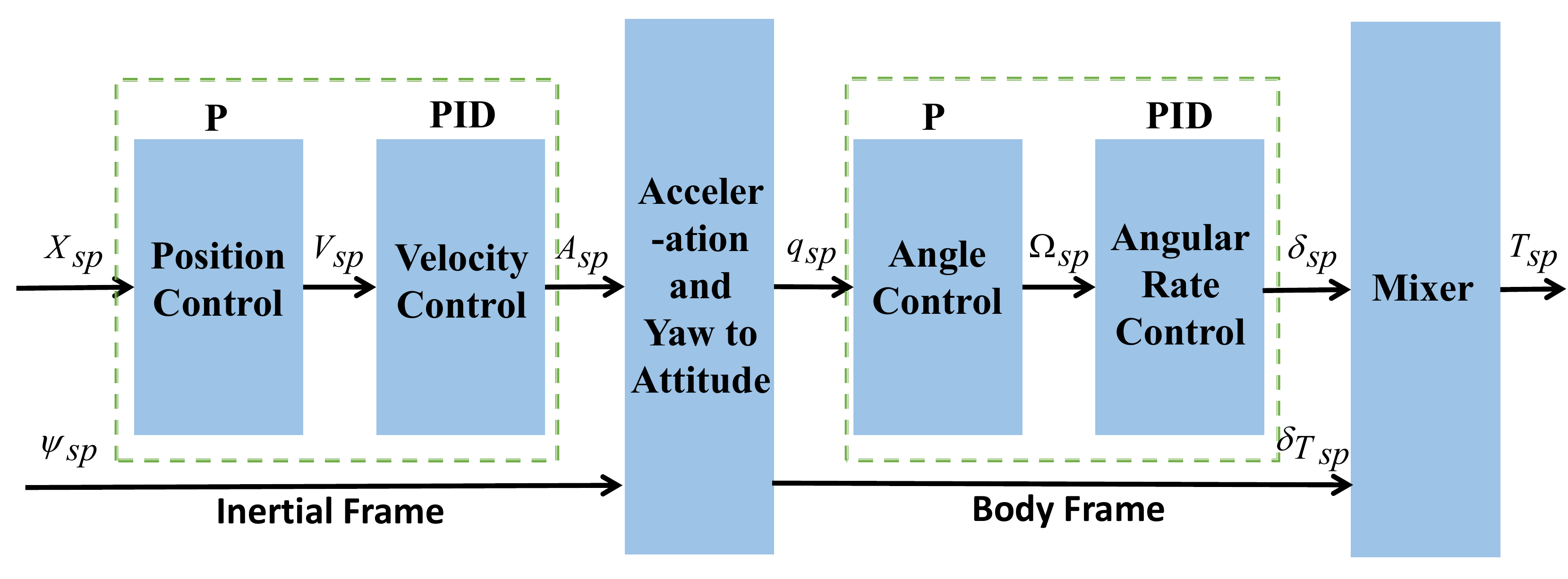}
    \caption{PX4 cascaded PID controller.}
    \label{fig:PX4_controller}
\end{figure}

A standard cascaded control architecture is recommended in PX4 for multirotor and the controllers are a mix of P and PID controllers, as shown in Fig.~\ref{fig:PX4_controller}. The controller consists of position control and attitude control. Inside the controller, the outer loop is for position (angle) control and the inner loop is for acceleration (angular velocity) control. The inner loop responds faster than the outer loop and acts directly on the motor; hence the inner loop parameters are vital when adjusting the controller parameters. In the offboard mode of PX4, position, velocity, angle and angular velocity can be controlled separately.

\section{DETECTION AND LOCALIZATION OF LANDING PLATFORM}

In this section, we introduce the detection of the landing platform and the calculation of the relative pose relationship between the UAV and the UGV, which is crucial to achieving a reliable landing.

\subsection{Pinhole Camera Calibration} 
For the purpose of mapping objects in three-dimensional space to two-dimensional camera space, we achieve pinhole camera calibration~\cite{zhang2000flexible} to establish the geometric model of a specific camera. The geometric model parameters are called camera parameters, including intrinsic and extrinsic parameters.

\begin{equation}
Z\left[ \begin{array}{c}
	u\\
	v\\
	1\\
\end{array} \right] =\left( \begin{matrix}
	f_x&		\gamma&		c_x\\
	0&		f_y&		c_y\\
	0&		0&		1\\
\end{matrix} \right) \left[ \begin{matrix}
	R_{3\times 3}&		t_{3\times 1}\\
\end{matrix} \right] \left[ \begin{array}{c}
	X\\
	Y\\
	Z\\
	1\\
\end{array} \right] 
\end{equation}

\subsection{Landing Pad Detection}

\begin{figure}[htb]
    \centering
    \includegraphics[width=0.3\columnwidth]{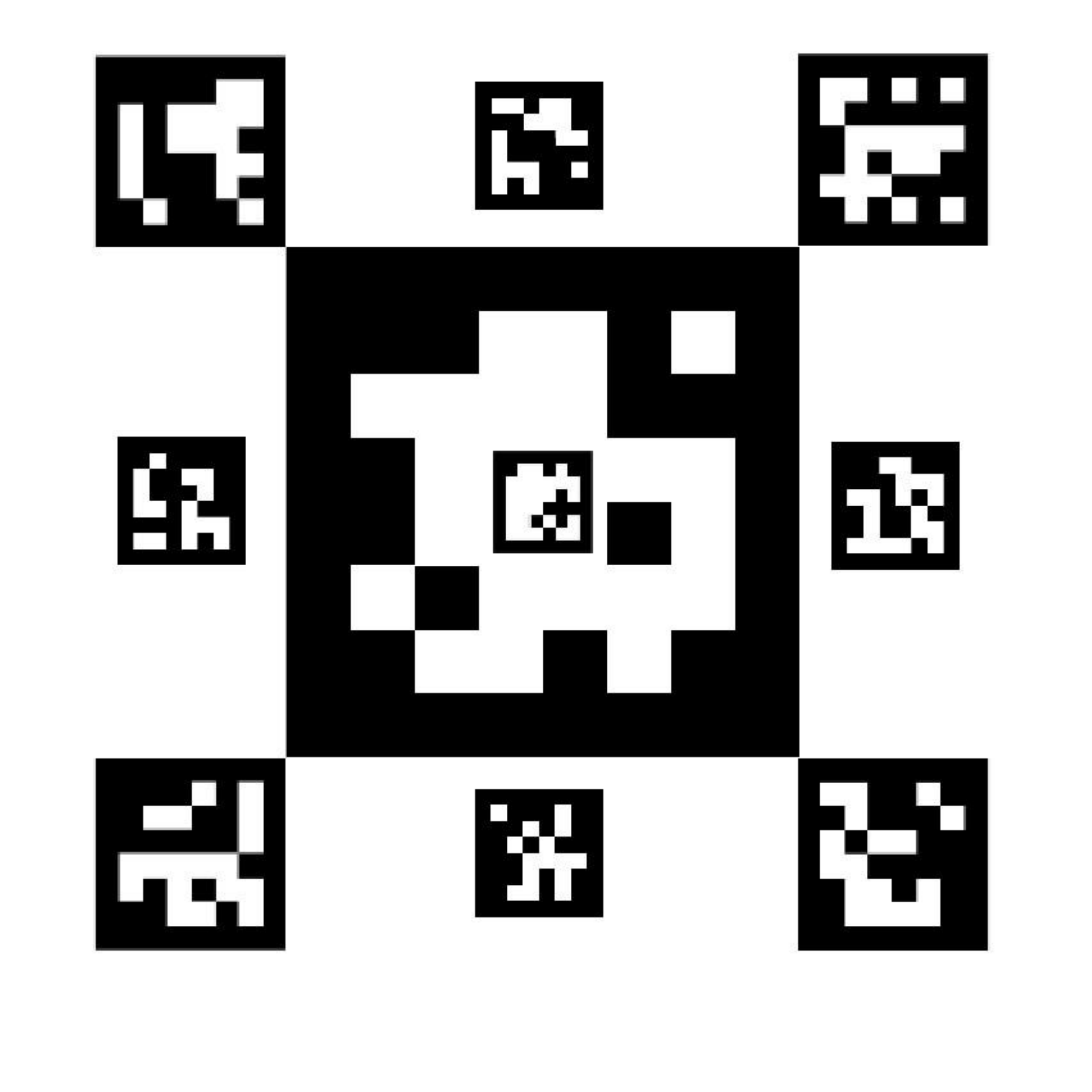}
    \caption{AruCo marker-based landing pad.}
    \label{fig:aruco}
\end{figure}

In this paper, we use the AruCo squared fiducial marker library~\cite{romero2018speeded} for target detection. We implement the detection of markers in OpenCV \cite{bradski2008learning}, which includes three steps: using adaptive thresholding to obtain borders, using polygonal approximation and removing too close rectangles, performing marker identification. The AruCo landing pad design is inspired by Qi \emph{et al.}~\cite{qi2019autonomous}, which consists of four different sizes and ten kinds of markers, as shown in Fig.~\ref{fig:aruco}. These markers ensure that the moving platform can be seen near, far and sideways. The actual scale of each AruCo marker and its accurate relative relationship with other surrounding markers are significant. For a more accurate landing, we define and test markers' maximum detectable and active distance ranges, as shown in Tab.~\ref{tab:marker detection}. The maximum distance range and offset means the maximum \emph{x, y, z} distance of the UAV from the 
center of a specific marker when the UAV can detect that specific marker. The active range is a subset of the maximum range, which means the distance at which the marker at a specific location plays a major role in calculating relative pose.

\begin{table}[H]
    \setlength{\abovecaptionskip}{-0.5cm}  
    \setlength{\belowcaptionskip}{0cm} 
    \caption{Marker detection range}
    \label{table_example}
    \begin{center}
    \resizebox{.5\textwidth}{!}{
        \renewcommand\arraystretch{1.5}
        \begin{tabular}{c|cc}
        \bottomrule \hline
            Marker type No. & Max detectable z-distance range & Max (x,y) offset\\
        \hline
            43 (2.5 cm) & (Land,0.15) m & (0.15,0.15) m\\
            5-8 (6.4 cm) & (Land,0.50) m & (0.39,0.39) m\\
            1-4 (9.5 cm) & (Land,1.15) m & (0.90,0.90) m\\
            68 (25.7 cm) & (Land,3.00) m & (1.42,1.42) m\\
        \bottomrule \hline
            Marker type No. & Active z-distance range & Active (x,y) offset\\
            \hline
            43 (2.5 cm) & (Land,0.15) m & (0.15,0.15) m\\
            5-8 (6.4 cm) & (0.20,0.30) m & (0.00,0.20-0.39) m\\
            1-4 (9.5 cm) & (0.40,1.00) m & (0.70,0.70) m\\
            68 (25.7 cm) & (1.00,3.00) m & -\\
        \hline \toprule 
        \end{tabular}
    }
    \end{center}
    \label{tab:marker detection}
\end{table}
\vspace{-0.5cm} 

\subsection{3D Pose Estimation}

Knowing n reference 3D points and corresponding 2D projections, we can construct the Perspective-n-Point (PNP) problem to estimate the camera's pose relative to the moving platform. When solving the PNP problem, similar to~\cite{qi2019autonomous}, we use the nonlinear optimization method \emph{Bundle Adjustment} to minimize the re-projection error and then estimate the rotation and translation between the camera coordinates and the marker coordinates. Through coordinate transformation, the coordinates in the landing pad system $M_q=\left\{ m_x,m_y,m_z \right\}$ can be transformed into the camera coordinate system $C_q=\left\{ c_x,c_y,c_z \right\}$, body-fixed coordinate system $B_q$ and the world inertial system (ENU) $E_q$ successively. Finally, the pose estimation of the UAV relative to the landing platform is expressed as
\begin{equation}
P^b=R_{C}^{B}\left( \left( R_{M}^{C}P^m+t_{M}^{C} \right) +Offset^c \right) 
\end{equation}
where $P,R,t$ denotes position, rotation and translation in different coordinate frames, respectively and $Offset^c$ is the known camera mount offset.

\section{PLANNING ALGORITHM AND LANDING STATE MACHINE}

Our paper uses a gradient-based local motion planning method~\cite{zhou2020ego}~\cite{zhou2021ego}, which does not need to construct the Euclidean Signed Distance Field (ESDF) in advance and therefore greatly reduces the planning time. Constructing the ESDF field takes not only plenty of time but also the trajectory calculated by the ESDF may fall into a local minimum. The method includes three steps: trajectory initialization, gradient-based trajectory optimization, time re-assignment and trajectory refinement.

\subsection{Front-end B-Spline Initialization}

In the initialization phase, a uniform B-Spline curve $\phi$ that satisfies the final constraints but does not consider obstacles is generated. An estimate of the collision force is performed and for each segment of the track where collision is detected, the A* algorithm
is used to generate a collision-free path $\varGamma$. Then for each control point $Q_i$ of
the collision line, an anchor point $P_{ij}$ is generated on the obstacle surface and the distance from $Q_i$ to the $jth$ obstacle is set as: $d_{ij}=\left( Q_i-P_{ij} \right) V_{ij}$, where $V_{ij}$ are unit vectors from $Q_i$ to $P_{ij}$.

\subsection{Back-end Trajectory Optimization}

The B-spline curve of the front-end part is uniquely determined by the degree $p_b$, $N_c$ control points $\left\{ Q_1,Q_2,\cdots ,Q_N \right\}$, and vector $\left[ t_1,t_2,\cdots ,t_M \right]$. The optimization problem is modeled as $min_QJ=\lambda _sJ_s+\lambda _cJ_c+\lambda _dJ_d$. In the formula, $J_s,J_c,J_d$ represents smoothness term, collision term, dynamic feasibility term, respectively and $\lambda _s,\lambda _c,\lambda _f$ is the corresponding coefficient. According to the convex hull property, the smoothness cost $J_s$ and feasibility cost $J_d$ is set as
\begin{equation}
J_s=\sum_{i=1}^{N_c-2}{\lVert A_i \rVert _{2}^{2}}+\sum_{i=1}^{N_c-3}{\lVert J_i \rVert _{2}^{2}}
\end{equation}
\begin{equation}
J_d=\sum_{i=1}^{N_c-1}{\omega _vF\left( V_i \right)}+\sum_{i=1}^{N_c-2}{\omega _aF\left( A_i \right)}+\sum_{i=1}^{N_c-3}{\omega _jF\left( J_i \right)}
\end{equation}
where $A_i$ is acceleration, $J_i$ is jerk, $\omega _{i,i=v,a,j}$ are weights and $F\left( \cdot \right)$ is a two-order continuously differentiable function of control points. Collision cost $J_c$ pushes control points until $d_{ij}>s_f
$(safe distance) and it is defined as
\begin{equation}
J_c=\sum_{i=1}^{N_c}{j_c\left( Q_i \right)}
\end{equation}
where $j_c\left( \cdot \right) $ is also a two-order continuously differentiable function.

\subsection{Time Re-assignment and Trajectory Refinement}

An additional time re-assignment step is necessary to avoid aggressive motion and ensure that the trajectory satisfies the kinodynamic constraints. Then, an curve fitting method is presented to make the refined trajectory $\varPhi _f$ maintain an almost identical shape to the original trajectory $\varPhi _s$. After obtaining the initial value of $\varPhi _f$, the refined optimization problem is defined as $min_QJ=\lambda _sJ_s+\lambda _cJ_c+\lambda _fJ_f$ and the third term $\lambda _fJ_f$ is the curve fitting term, which is defined as
\begin{equation}
J_f=\int_0^1{\left( \frac{d_{\alpha}\left( \alpha \right) ^2}{a^2}+\frac{d_r\left( \alpha \right) ^2}{b^2} \right)}d\alpha 
\end{equation}
where $d_{\alpha}\left( \alpha \right)$ and $d_r\left( \alpha \right)$ represents axial and radial displacement of two curves, respectively. $a$ is set to 20 and $b$ is set to 1. 


\subsection{Landing State Machine}

\begin{figure}[htb]
    \centering
    \includegraphics[width=1.0\columnwidth]{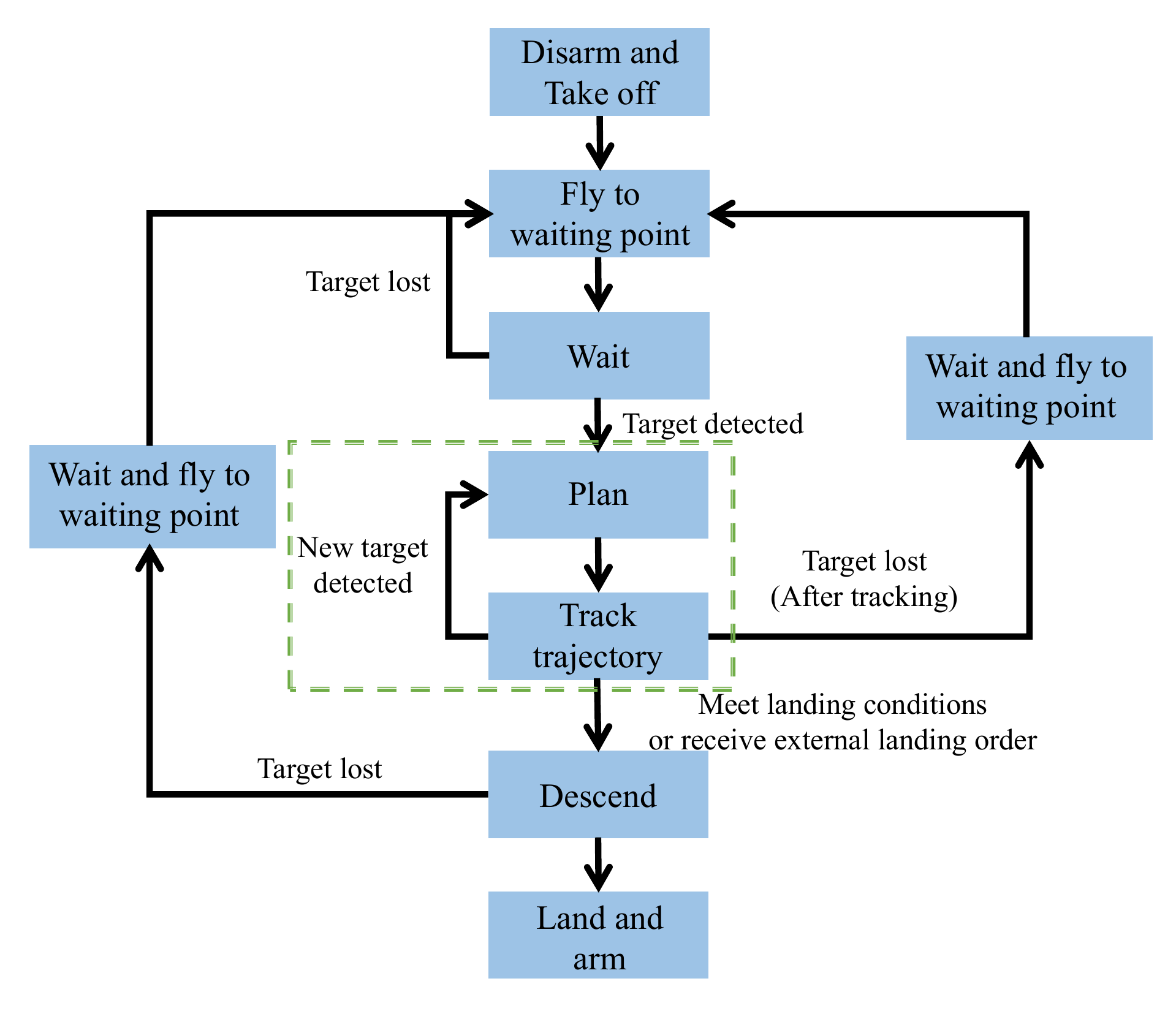}
    \caption{Landing state machine.}
    \label{fig:state_machine}
\end{figure}

The UAV's autonomous take-off and landing task is driven by an autonomous state machine, as shown in Fig.~\ref{fig:state_machine}. The main steps are as follows: UAV takes off and flies to a preset point; UAV hovers and waits for the landing target to appear; UAV sees the landing point and calculates the relative pose; UAV's planner plans a reference trajectory in real-time; UAV's cascade PID controller tracks the generated trajectory; UAV meets the landing conditions, descends, and completes the landing task. It is worth noting that because the target point is moving, the UAV usually does not entirely execute the trajectory given by the planner, and will instead execute a new reference trajectory.

\vspace{-0.5cm} 

\section{EXPERIMENTS AND RESULTS}

In this section, we test our method in both simulation and real-world experimental environments.

\subsection{Simulation Experiments}

We conduct simulation experiments on the ROS/Gazebo platform named \emph{Prometheus}\footnote{Qi, Y., Jin, R., Jiang, T., and Li, C. (2020). Prometheus, Amov lab. Retrieved October 20, 2020, https://github.com/amov-lab.} and we also use part of the ROS packages provided by \emph{Prometheus} in real-world experiments.

\begin{figure}[htb]
    \centering
    \includegraphics[width=0.8\columnwidth]{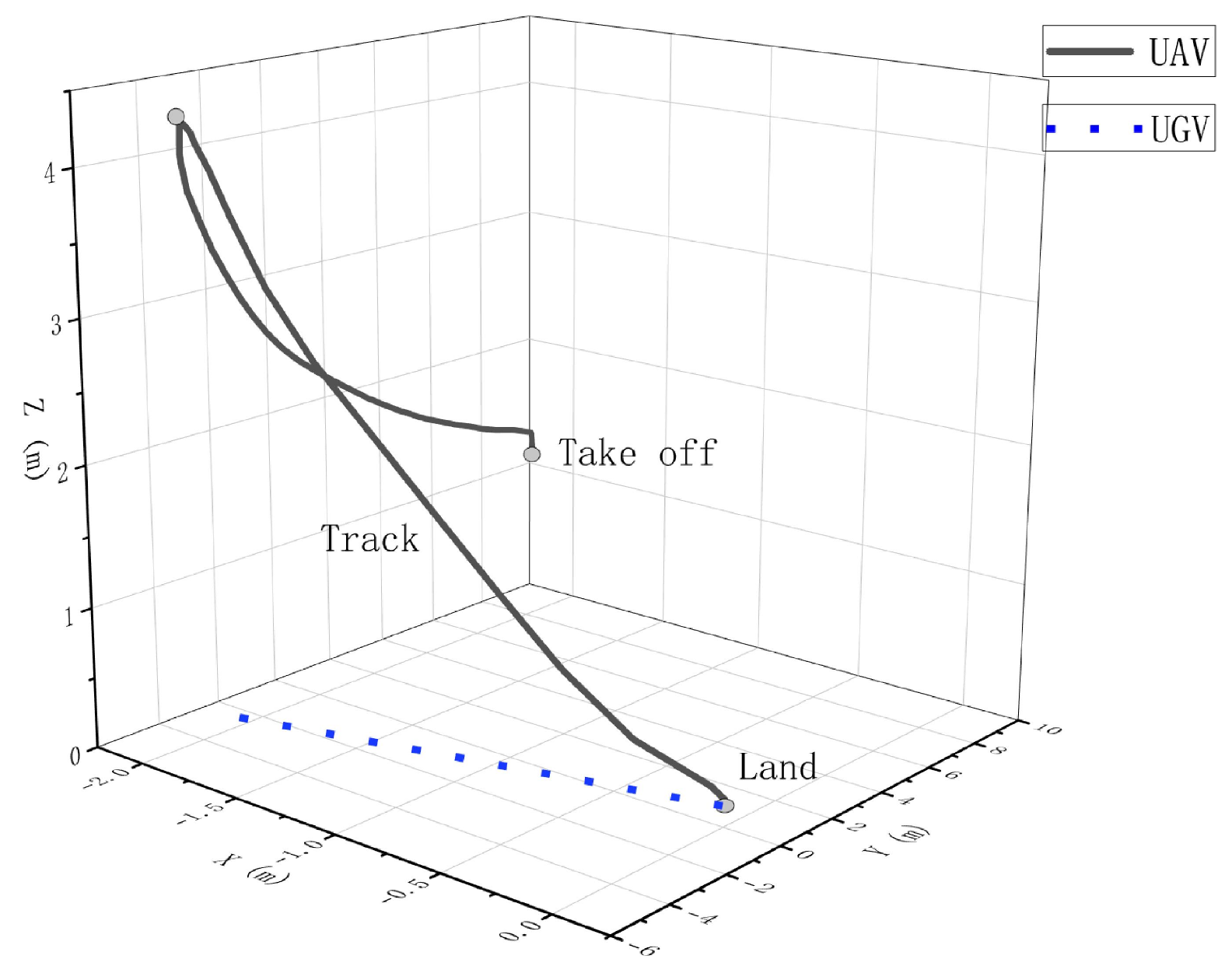}
    \caption{Simulation trajectory. Target speed: 2.8 km/h.}
    \label{fig:simu_traj}
\end{figure}

The UAV takes off to the pre-set point $\left( -4.0,-2.0,4.5 \right)$ and waits to detect the landing platform. After detecting the UGV, the UAV continuously plans the trajectory and tracks it. Then UAV descends and when the height from the landing point is less than 0.6m, it starts to land and completes the landing. The trajectory of UAV is shown in Fig.~\ref{fig:simu_traj} and main flight data is shown in Tab.~\ref{tab:flight data}.

\begin{table}[]
    \setlength{\abovecaptionskip}{-0.5cm}  
    \setlength{\belowcaptionskip}{0cm} 
    \caption{Key Flight Data}
    \label{table_example}
    \begin{center}
    \resizebox{.5\textwidth}{!}{
        \renewcommand\arraystretch{1.5}
        \begin{tabular}{c|ccccccc}
        \bottomrule \hline
                 Experiment & Distance & Target Speed & UAV Max Speed & UAV Average Speed & Planning Time\\
        \hline
            Simulation & 20 m & 2.8 km/h & 4.1 km/h & 0.8 km/h & 4.75 ms\\
            Indoor 1 & 5.2 m & 0.0 km/h & 4.8 km/h & 0.4 km/h & 0.83 ms\\
            Indoor 2 & 3.7 m & 0.8 km/h & 2.1 km/h & 0.3 km/h & 1.60 ms\\
            Outdoor 1 & 12.9 m & 2.16 km/h & 6.6 km/h & 2.7 km/h & 3.16 ms\\
            Outdoor 2 & 14.3 m & 3.24 km/h & 8.2 km/h & 3.3 km/h & 3.49 ms\\
        \hline \toprule 
        \end{tabular}
    }
    \end{center}
    \label{tab:flight data}
\end{table}

\subsection{Real-world Experiments}

\subsubsection{Indoor Environment}

\begin{figure}[htb]
    \centering
    \includegraphics[width=0.8\columnwidth]{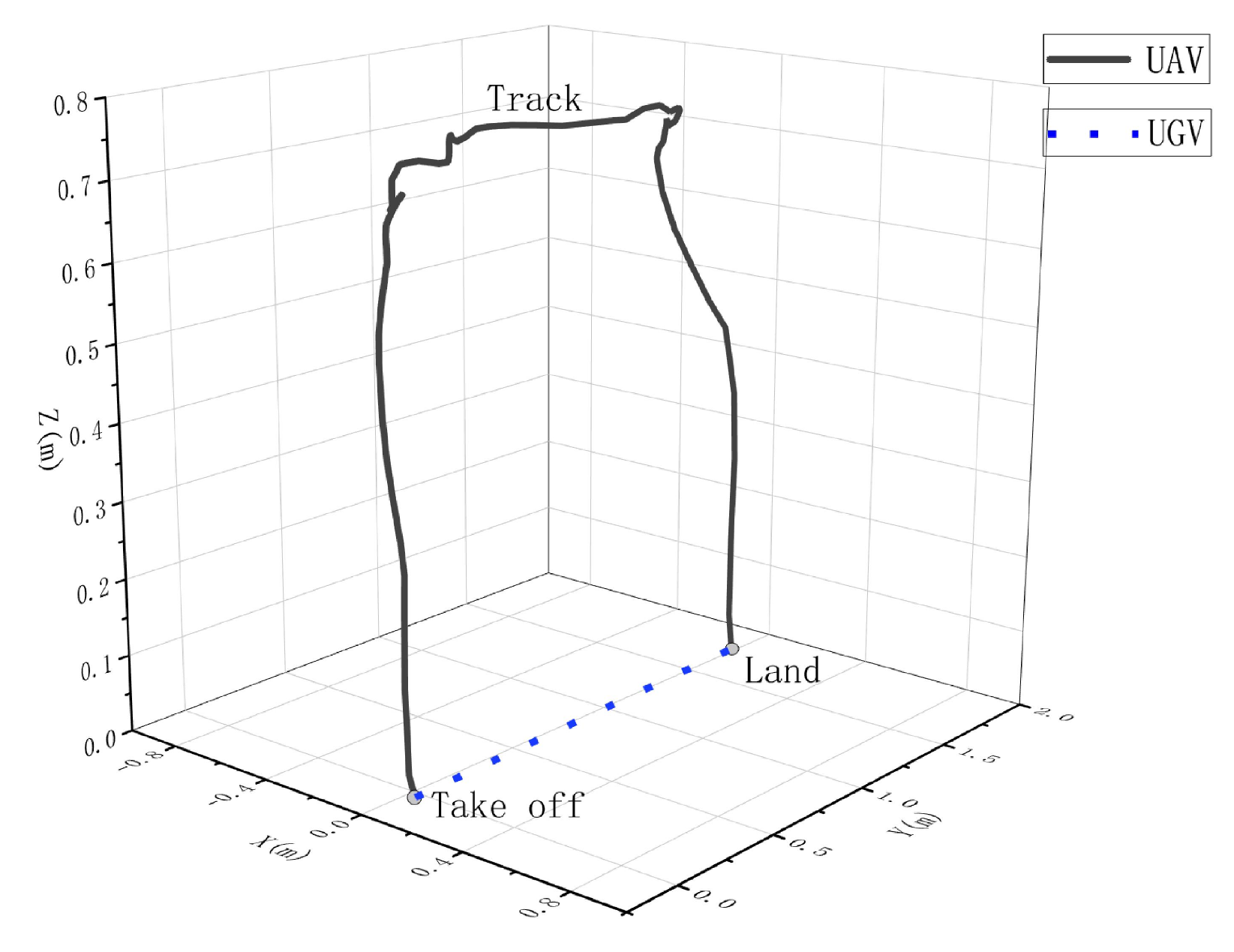}
    \caption{Indoor experiment trajectory. Target speed: 0.8 km/h.}
    \label{fig:indoor}
\end{figure}

We first conduct experiments on static and dynamic targets in indoor environments. The UAV can accurately land on the target within the field of view. In a moving target experiment, the UGV moves linearly in the positive direction of the y-axis at a speed of 0.8 km/h. After the UAV takes off to the target point $\left(0.0,0.0,0.75 \right)$, it starts to follow, and when the landing conditions are satisfied (tracking distance over 2.5m or external landing command), it begins to descend to 0.2m, and then quickly lands. The trajectory of the UAV and UGV is shown in Fig.~\ref{fig:indoor} and the experimental data is shown in Tab.~\ref{tab:flight data}.


\subsubsection{Outdoor Environment with Wind Speed}

\begin{figure}[htb]
    \centering
    \includegraphics[width=0.8\columnwidth]{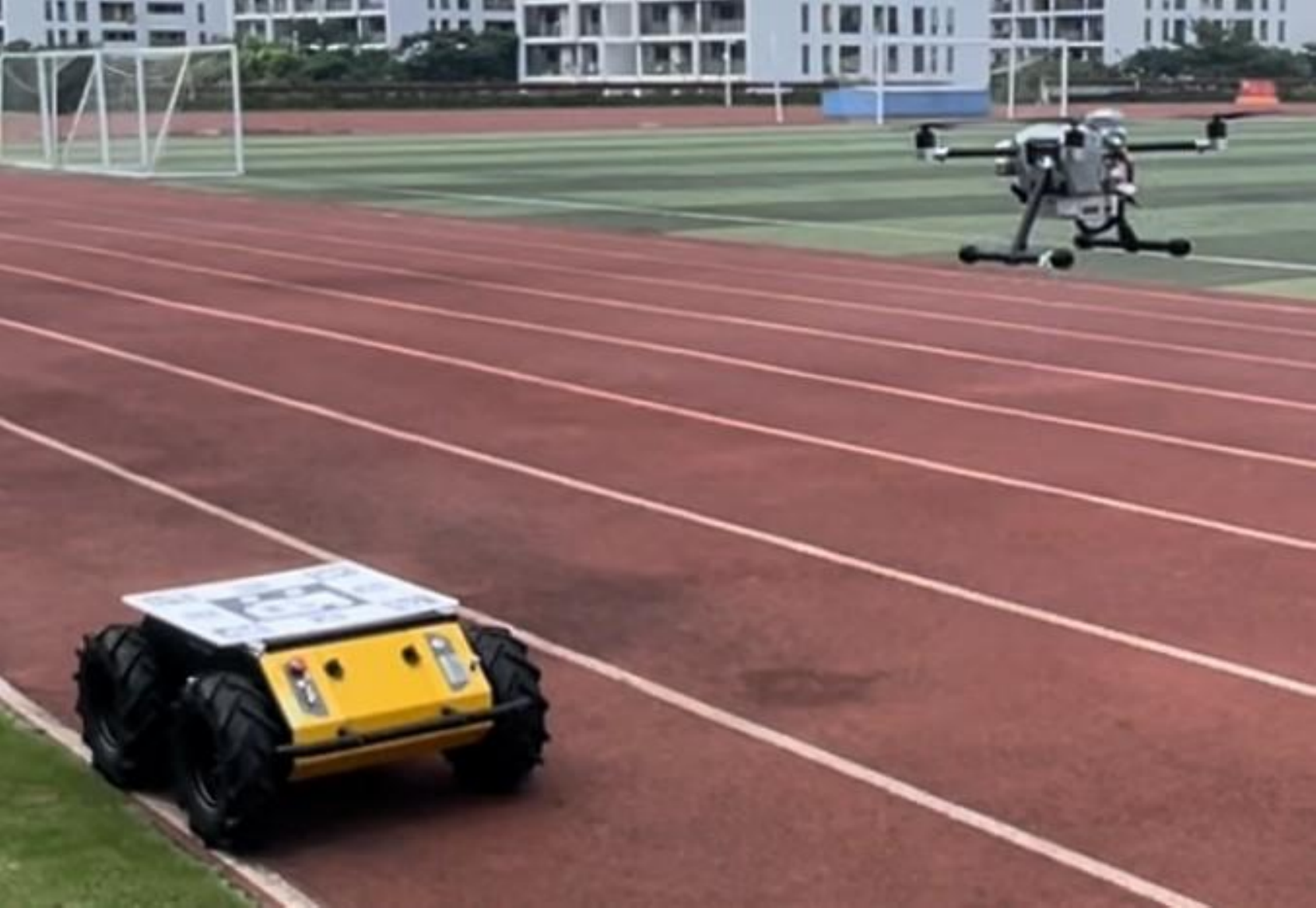}
    \caption{Outdoor environment.}
    \label{fig:environment}
\end{figure}

\begin{figure}[htb]
    \centering
    \includegraphics[width=0.9\columnwidth]{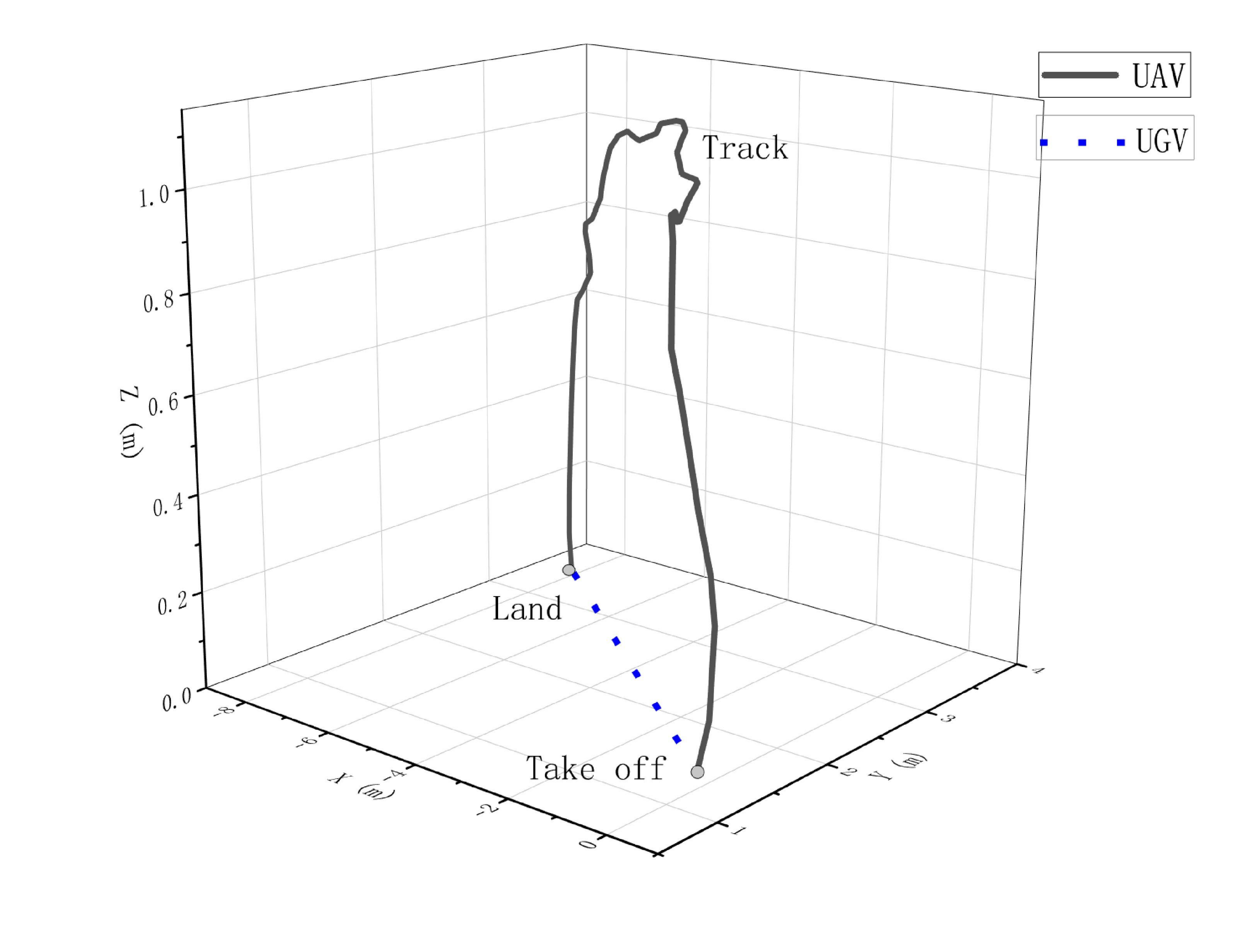}
    \caption{Outdoor experiment trajectory. Target speed: 3.24 km/h.}
    \label{fig:outdoor}
\end{figure}

We then conduct dynamic target landing experiments in an outdoor environment with wind speed, as shown in Fig.~\ref{fig:environment}. 
Similar to the indoor experiment, the UGV moves at a linear speed of 2.16 km/h and 3.24 km/h, respectively; the UAV first takes off to a preset point $\left(0.0,0.0,1.0 \right)$, then follows, and when an external landing command is received, the UAV quickly lands. The trajectory of the UAV-UGV system is shown in Fig.~\ref{fig:outdoor} and the critical data is also shown in Tab.~\ref{tab:flight data}.

\subsubsection{Trajectory Analysis}
It can be seen from the trajectory graph that the trajectory in the simulation is relatively smooth. Although the motion planner gives a smooth reference trajectory, the actual flight trajectory of the UAV in the real situation slightly oscillates. This is because the robot's actuators, including motors, have uncertainties from control noise, wear, and mechanical failures~\cite{thrun2002probabilistic}.

\subsection{Obstacle Avoidance Demonstration}

\begin{figure}[htb]
    \centering
    \includegraphics[width=0.99\columnwidth]{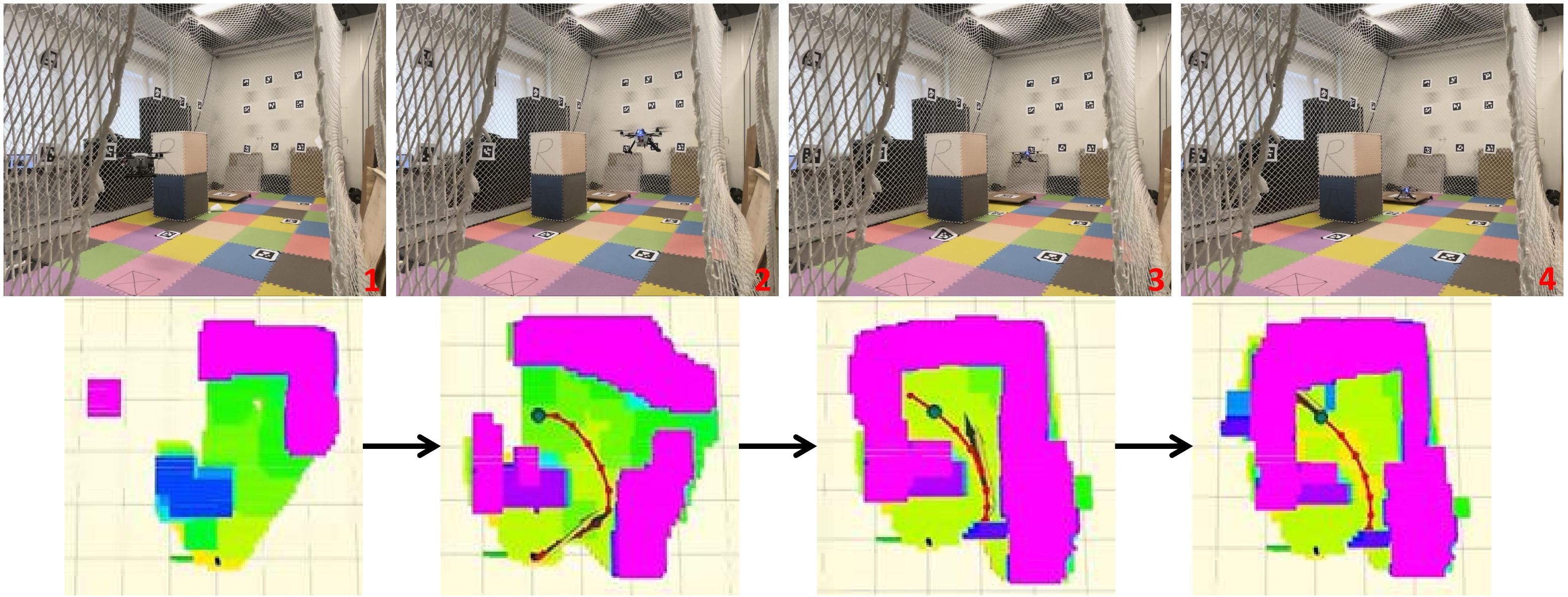}
    \caption{Obstacle avoidance and trajectory update demonstration.}
    \label{fig:obstacle_avoid}
\end{figure}

We use the local motion planner in~\cite{zhou2020ego} and provide the UAV with collision-free, smooth and feasible trajectories. In the experiment, there is a $60cm\times 120cm\times 120cm$ obstacle in front of the UAV. The UAV bypasses the obstacle, and then sees and lands on the platform, as shown in Fig.~\ref{fig:obstacle_avoid}. 

Crucial planning data is shown in Tab.~\ref{tab:obstacle avoidance}, where the \emph{Obstacle Density} is the proportion of obstacles per square meter.

\begin{table}[H]
    \setlength{\abovecaptionskip}{-0.5cm}  
    \setlength{\belowcaptionskip}{0cm} 
    \caption{Obstacle Avoidance}
    \label{table_example}
    \begin{center}
    \resizebox{.5\textwidth}{!}{
        \renewcommand\arraystretch{1.5}
        \begin{tabular}{ccccc}
        \bottomrule \hline
                Flight Time & Flight Distance & Obstacle Density & Planning Time & Re-planning Times\\
            58 s & 10.2 m & 0.11$/m^2$ & 2.40 ms & 5 times\\
        \hline \toprule 
        \end{tabular}
    }
    \end{center}
    \label{tab:obstacle avoidance}
\end{table}

\subsection{Parameters Setting Analysis}

We analyze and tune the prime parameters of the motion planner, as shown in Tab.~\ref{tab:planner parameters}. \emph{Resolution} is the resolution of the grid map of the surrounding environment built from the depth information and \emph{Obstacles Inflation} is the relative size of the inflation of the obstacles. $\lambda$ is the coefficient of each loss item in the loss function described in IV. For the sake of safety, the collision $\lambda _{collision}$ in our paper is set relatively large.

\begin{table}[H]
    \setlength{\abovecaptionskip}{-0.5cm}  
    \setlength{\belowcaptionskip}{0cm} 
    \caption{Planner Parameters}
    \label{table_example}
    \begin{center}
    \resizebox{.5\textwidth}{!}{
        \renewcommand\arraystretch{1.5}
        \begin{tabular}{cccccc}
        \bottomrule \hline
                Resolution & Obstacles Inflation & $\lambda _{smooth}$ & $\lambda _{collision}$ & $\lambda _{feasibility}$ & $\lambda _{fitness}$\\
            0.15 & 0.299 & 1.0 & 8.5 & 0.1 & 1.0 \\
        \hline \toprule 
        \end{tabular}
    }
    \end{center}
    \label{tab:planner parameters}
\end{table}

\subsection{Comparison Results}

We compare our method with previous work in autonomous landing to verify the effectiveness of our method. The relevant comparison results are shown in Tab.~\ref{tab:comparison results}. Compared with the previous methods, our system can realize the autonomous take-off and landing of a quadrotor on a UGV platform in different scenarios without external positioning and can avoid static obstacles simultaneously.

\begin{table}[H]
    \setlength{\abovecaptionskip}{-0.5cm}  
    \setlength{\belowcaptionskip}{0cm} 
    \caption{Comparison Results with Related Works}
    \label{table_example}
    \begin{center}
    \resizebox{.5\textwidth}{!}{
    	\renewcommand\arraystretch{1.5}
        \begin{tabular}{c|cccc}
        \bottomrule \hline
            Work & Target Speed & Obstacle Avoidance & External Localization & Environment \\
        \hline
            Sani \emph{et al.} & 0.0 km/h & No & No need & Indoor\\
            Lee \emph{et al.} & 0.25 km/h & No & No need & Indoor\\
            Mellinger \emph{et al.} & 0.0 km/h & No & Need & Indoor\\
            Daly \emph{et al.} & 3.6 km/h & No & Need & Outdoor\\
            Ours & 3.24 km/h & Yes & No need & Indoor/Outdoor\\
        \hline \toprule 
        \end{tabular}
    }
    \end{center}
    \label{tab:comparison results}
\end{table}

\section{CONCLUSIONS AND FUTURE WORK}

In this paper, we present a quadrotor system in the above sections. In order to detect the target quickly and accurately, we design an AruCo-based landing pad system. We utilize a gradient-based local motion planner to rapidly generate collision-free, smooth, and kinodynamic feasible reference trajectories. We then present an automatic state machine to achieve full autonomy in take-off and landing tasks. Experiments show that our system could realize the autonomous landing task on a mobile platform in real outdoor environments. 

In the future, we will focus on more high-speed and more uncertain moving target landing tasks, while focusing on improving the accuracy of landing.

\section*{ACKNOWLEDGMENT}

We appreciate Pengqin Wang and Delong Zhu for their constructive advice.










{\small
\bibliographystyle{ieeetr}
\bibliography{root}
}

\end{document}